\PassOptionsToPackage{table,dvipsnames}{xcolor}
\documentclass[10pt,twocolumn,letterpaper]{article}
\usepackage[final]{iccv}      
\usepackage{algorithm}
\usepackage{algorithmic}
\usepackage{multirow}
\usepackage{threeparttable}
\usepackage{colortbl}
\usepackage{etoolbox}
\usepackage{booktabs}
\usepackage{caption}
\usepackage[accsupp]{axessibility}
\definecolor{iccvblue}{rgb}{0.21,0.49,0.74}
\usepackage[pdftex,pagebackref,breaklinks,colorlinks,allcolors=iccvblue]{hyperref}
\usepackage[table]{xcolor}
\def\paperID{9956} 
\def\confName{ICCV} 
\def\confYear{2025}
\title{From Objects to Events: Unlocking Complex Visual Understanding in \\ Object Detectors via LLM-guided Symbolic Reasoning
}
\author{
Yuhui Zeng$^1$\thanks{Equal contribution.} , Haoxiang Wu$^1$\footnotemark[1] , Wenjie Nie$^1$, Guangyao Chen$^2$\thanks{Corresponding authors.} ,
Xiawu Zheng$^1$\footnotemark[2] ,\\
Yunhang Shen$^3$, Jun Peng$^1$, Yonghong Tian$^2$, Rongrong Ji$^1$  \\
$^1$Key Laboratory of Multimedia Trusted Perception and Efficient Computing, \\
Ministry of Education of China, Xiamen University \\
$^2$ Peking University \quad $^3$Tencent Youtu Lab \\
}
\begin{document}
\maketitle
\renewcommand*{\thefootnote}{\fnsymbol{footnote}}
\newcounter{ct}
\begin{abstract}
Current object detectors excel at entity localization and classification, yet exhibit inherent limitations in event recognition capabilities. 
This deficiency arises from their architecture's emphasis on discrete object identification rather than modeling the compositional reasoning, inter-object correlations, and contextual semantics essential for comprehensive event understanding. 
To address this challenge, we present a novel framework that expands the capability of standard object detectors beyond mere object recognition to complex event understanding through LLM-guided symbolic reasoning. 
%
Our key innovation lies in bridging the semantic gap between object detection and event understanding without requiring expensive task-specific training. 
The proposed plug-and-play framework interfaces with any open-vocabulary detector while extending their inherent capabilities across architectures. 
At its core, our approach combines (\romannumeral1) a symbolic regression mechanism exploring relationship patterns among detected entities and (\romannumeral2) a LLM-guided strategy guiding the search toward meaningful expressions. 
These discovered symbolic rules transform low-level visual perception into interpretable event understanding, providing a transparent reasoning path from objects to events with strong transferability across domains.
We compared our training-free framework against specialized event recognition systems across diverse application domains. 
Experiments demonstrate that our framework enhances multiple object detector architectures to recognize complex events such as illegal fishing activities ($\textbf{75\%}$ AUROC, $\textbf{+8.36\%}$ improvement), construction safety violations ($\textbf{+15.77\%}$), and abnormal crowd behaviors ($\textbf{+23.16\%}$). 
Code is available at \href{https://github.com/MAC-AutoML/SymbolicDet}{here}.
%
\end{abstract}
%
%
%
\vfill

\section{Introduction}
\label{sec:intro}
\begin{figure}[!t]
    \centering
    \setlength{\abovecaptionskip}{0cm}
    \includegraphics[width=1\linewidth]{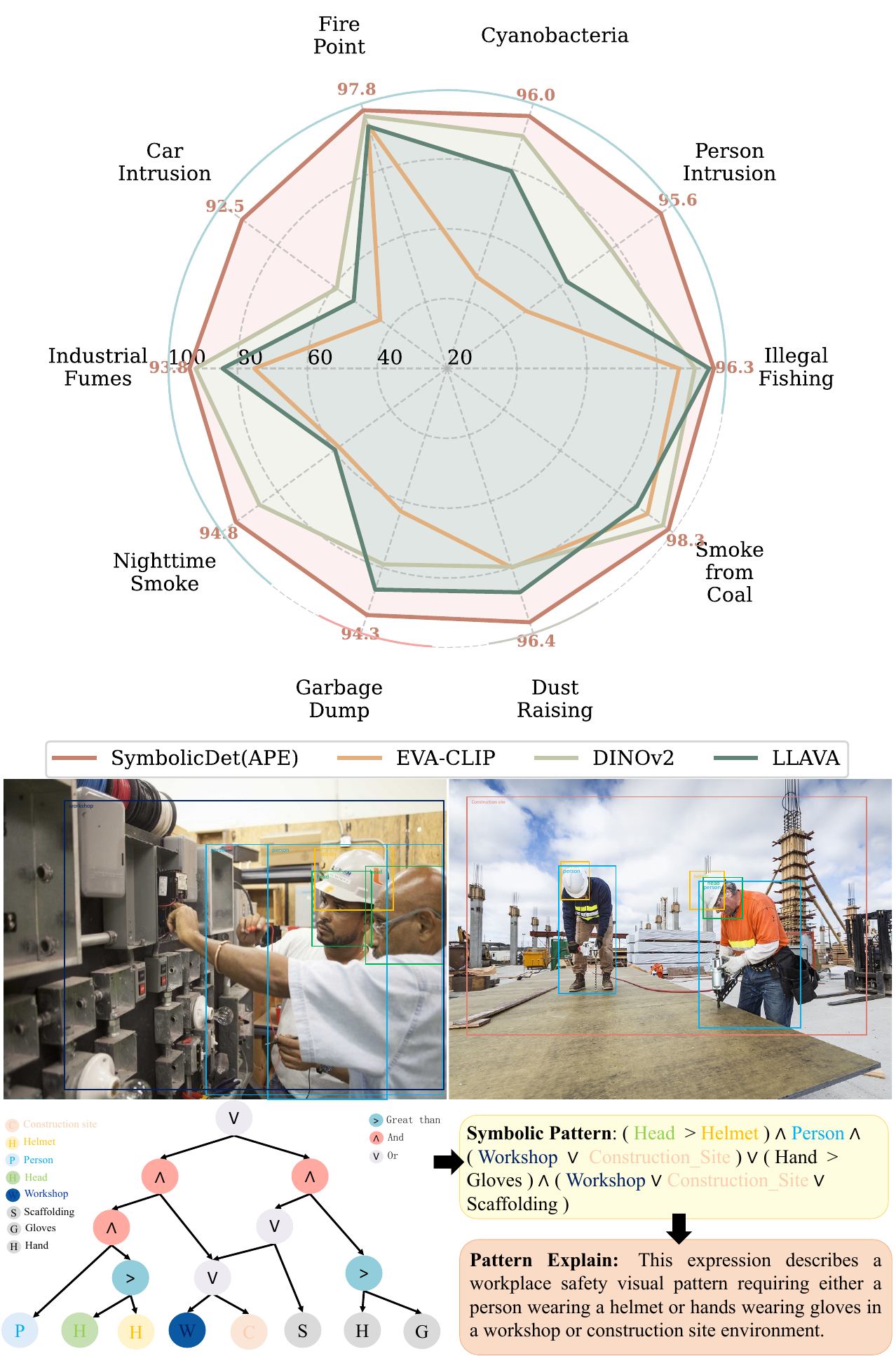}
    \caption{The radar chart at the top illustrates the comparative performance of various models (SymbolicDet(APE \cite{shen2024aligning}), EVA-CLIP \cite{sun2023eva}, DINOv2 \cite{oquab2023dinov2}, LLAVA \cite{liu2023visual}) across different event detection scenarios. The bottom section provides a visual representation of workplace safety patterns identified through our SymbolicDet framework, showcasing specific conditions like helmet and glove usage in workshop or construction site environments.}
    \label{fig:composite}
    \vspace{-1.0cm}
\end{figure}
Object detection has become a cornerstone of computer vision, enabling machines to identify and locate entities within visual scenes with remarkable accuracy \cite{girshick2014rich, girshick2015fast, ren2015faster, redmon2016you,cai2018cascade,carion2020end,zhu2020deformable,zhang2022dino,jia2023detrs,liu2022dab,meng2021conditional}. State-of-the-art detectors can now recognize thousands of object categories with remarkable precision, transforming applications across autonomous driving, surveillance, industrial inspection, and content analysis. However, despite these advances, a fundamental limitation persists: while modern detectors excel at answering ``what is present" in an image, they struggle with understanding ``what is happening" --- the relationships, interactions, and events occurring between detected entities. \par
Consider a coastal surveillance scenario where a detector identifies ``persons" and ``fishing rods" with high confidence. Despite perfect detection, the system fails to recognize the critical event of ``illegal multi-rod fishing" --- where a single individual operates multiple fishing rods, violating conservation regulations. Similarly, in construction site monitoring, a standard detector might accurately identify workers, equipment, and safety gear, yet remain incapable of recognizing the crucial safety violation where a worker operates machinery without proper protective equipment.
These limitations stem from the architectural focus of object detectors on identifying discrete entities rather than modeling the compositional logic, relational semantics, and contextual dependencies that define meaningful events. Without the ability to understand ``what is happening" beyond ``what exists," even the most accurate detection systems fall short in scenarios requiring nuanced interpretation of object relationships and contextual significance --- a fundamental barrier to deploying truly intelligent visual systems in complex real-world scenarios. \par
Traditionally, bridging this gap has required two unsatisfactory approaches. The first involves developing specialized event recognition systems trained on extensive labeled datasets for each target event, incurring prohibitive annotation costs and limiting generalizability \cite{gehrig2023recurrent, liu2023motion,li2023sodformer}. The second approach employs fine-tuning techniques to adapt object detectors to specific events, which sacrifices their general-purpose nature and still requires substantial task-specific data \cite{liu2024grounding,saito2022learning,gupta2022ow,zohar2023prob,li2022asynchronous,wang2023dual,zhang2022spiking}. These approaches not only demand significant resources but also typically yield black-box models that provide little insight into their reasoning process --- a critical limitation in safety-critical or regulated domains where interpretability is essential. \par
We present a fundamentally different approach that fundamentally reframes the problem: rather than developing specialized event recognition models, we propose to unlock the latent event recognition capabilities inherent in standard object detectors through the integration of LLM-guided symbolic reasoning. Our key insight is that standard object detectors already implicitly encode rich visual information that, when properly interpreted through symbolic reasoning, can reveal complex events and relationships. Rather than treating detectors as mere entity recognizers, we view them as sophisticated visual sensors whose outputs can be transformed into meaningful event understanding through interpretable logical reasoning. \par
Our framework, consists of three synergistic components. First, an open-vocabulary object detector extracts entity-level information. Second, a symbolic reasoning module discovers logical patterns among these entities through an evolutionary search process, generating human-readable expressions that capture complex relationships. Third, and most innovatively, we leverage Large Language Models (LLMs) to guide this symbolic search, infusing the process with rich world knowledge and semantic understanding that dramatically improves search efficiency and expression quality.
This approach offers several significant advantages over existing methods. First, it operates in a training-free manner, requiring no additional labeled data beyond what the underlying detector was trained on. Second, it maintains complete interpretability, with all event recognition decisions expressed as readable logical rules (as exemplified in Figure \ref{fig:composite}). Finally, our method is detector-agnostic, functioning as a plug-and-play enhancement layer that can augment any object detection system. \par
Through extensive experiments across multiple datasets, we demonstrate that our approach successfully enhances various detector architectures (APE \cite{shen2024aligning}, GLIP \cite{li2022grounded}, and YOLO-World \cite{cheng2024yolo}) to recognize complex events including illegal fishing activities, construction safety violations, and abnormal crowd behaviors. In the UCSD Ped2 benchmark \cite{wang2010anomaly}, our training-free approach achieves 98.7\% AUROC, approaching state-of-the-art performance of specialized, training-intensive methods (99.7\%), while providing fully transparent reasoning. For safety helmet compliance detection, our method improves recognition accuracy by 15.77\% without any domain-specific training.
%
The principal contributions of our work include:
\begin{itemize}
    \item We propose a novel framework that unlocks complex event understanding capabilities in standard object detectors through LLM-guided symbolic reasoning, without requiring additional training.
    \item We develop an efficient mechanism for discovering interpretable symbolic patterns from object detector, enabling transparent reasoning from object-level detections to event-level understanding.
    \item We introduce a structured LLM reasoning process that guides symbolic search, leveraging natural language understanding to discover meaningful patterns while dramatically improving search efficiency.
    \item We introduce the Helmet-Mac Dataset, a comprehensive resource containing 12,213 samples specifically designed for construction safety compliance detection, which we make publicly available to the research community.
\end{itemize}
%
Through these contributions, we not only enhance the practical utility of object detection systems but also advance our understanding of how compositional reasoning can bridge low-level perception and high-level event semantics. Our work represents a step toward visual AI systems that not only see objects but understand the meaningful events unfolding within visual scenes.
\section{Related work}
\label{sec:related}
\subsection{Visual Reasoning and Neuro-symbolic}
Our work extends beyond conventional object detection to enable reasoning about complex visual events, placing it within the broader visual reasoning paradigm. While we leverage object detector outputs as our foundation, we transform these into symbolic representations suitable for higher-order reasoning.
Visual reasoning research has progressed from simple object recognition to complex scene understanding requiring compositional analysis. Traditional approaches have relied on specialized architectures and extensive labeled datasets for specific reasoning tasks. Visual question answering systems interpret images through natural language questions \cite{hudson2019gqa,chen2024spatialvlm,khan2024consistency,ganz2024question,majumdar2024openeqa,chen2024vtqa}, while scene graph generation approaches identify object relationships to construct structured scene representations \cite{krishna2017visual,li2024pixels, wang2025scene,lin2022ru,zareian2020bridging}. However, these methods often lack interpretability or require task-specific training data.
Neuro-symbolic models offer a promising direction by combining neural networks' perceptual strengths with symbolic reasoning's interpretability and compositionality \cite{garcez2019neural,amizadeh2020neuro,mao2019neuro,shi2019explainable,yi2018neural, li2023neural}. These approaches typically extract symbolic representations from visual scenes using neural networks, then apply symbolic reasoning methods to these representations. 
Our framework advances this paradigm by implementing a novel neuro-symbolic approach where object detectors serve as the neural perception component while a symbolic reasoning layer guided by LLMs performs higher-level event understanding. Unlike traditional implementations requiring custom integration between components, our approach treats existing object detectors as modular perception units, maintaining interpretability while enabling flexible application across visual domains without task-specific training.
\subsection{LLMs for Visual Tasks}
Our approach uniquely positions LLMs as reasoning guides for symbolic search rather than for direct visual perception. This design allows us to leverage LLMs' rich world knowledge while maintaining a clear separation between perception (via object detectors) and reasoning (via interpretable symbolic operations).
Recent advances in LLMs have demonstrated impressive capabilities in visual understanding \cite{radford2021learning,liu2021swin,han2023llms,sun2023eva,zhang2022dino,liang2023open,zou2024segment,shen2024aligning,cheng2024yolo}. DetGPT \cite{pi2023detgpt} enables open-vocabulary object detection through reasoning, while Groundhog \cite{zhang2024groundhog} grounds LLMs to holistic segmentation tasks. Despite these advances, directly applying LLMs to visual reasoning presents challenges due to modality gaps and reasoning complexity.
Our framework addresses these challenges by using LLMs in their native text domain to guide symbolic pattern discovery over detector outputs. This approach maintains complete interpretability throughout the process --- a critical advantage over end-to-end black-box models. By separating perception from reasoning, we combine neural models' perceptual capabilities with symbolic reasoning's interpretability, enhanced by LLMs' semantic understanding, without requiring extensive multimodal training. Unlike automated prompt optimization that uses search to refine LLM inputs \cite{guo2023connecting,pan2023plum,xu2022gps}, we use the LLM to guide an external symbolic search, rather than treating the LLM as the direct target of optimization.
\subsection{Event Recognition and Understanding}
Our work extends into event recognition, where we enable complex visual understanding by reasoning about compositional relationships between detected entities.
Traditional event recognition approaches typically rely on specialized architectures trained on event-specific datasets \cite{lin2021complex,sakaino2023deepunseen,merlo2023automatic,fan2023flexible}. These methods often struggle with novel event types or complex scenarios requiring compositional reasoning. More recent approaches leveraging large-scale pretraining have improved generalization capabilities but often lack interpretability and explicit reasoning mechanisms.
Our framework addresses these limitations by enabling compositional reasoning over object detections to recognize complex events. Our framework addresses these limitations through compositional reasoning over object detections. By transforming detector outputs into symbolic representations, we enable LLM-guided search to identify specific patterns of object interactions characterizing complex events.   Unlike methods requiring extensive event-specific training data, our approach can leverage existing object detectors and LLMs' reasoning capabilities to understand diverse event types without additional visual training.
\section{Method}
\label{sec:method}
\begin{figure*}[t!]
    \centering
    \setlength{\abovecaptionskip}{0.1cm}
    \includegraphics[width=1\linewidth]{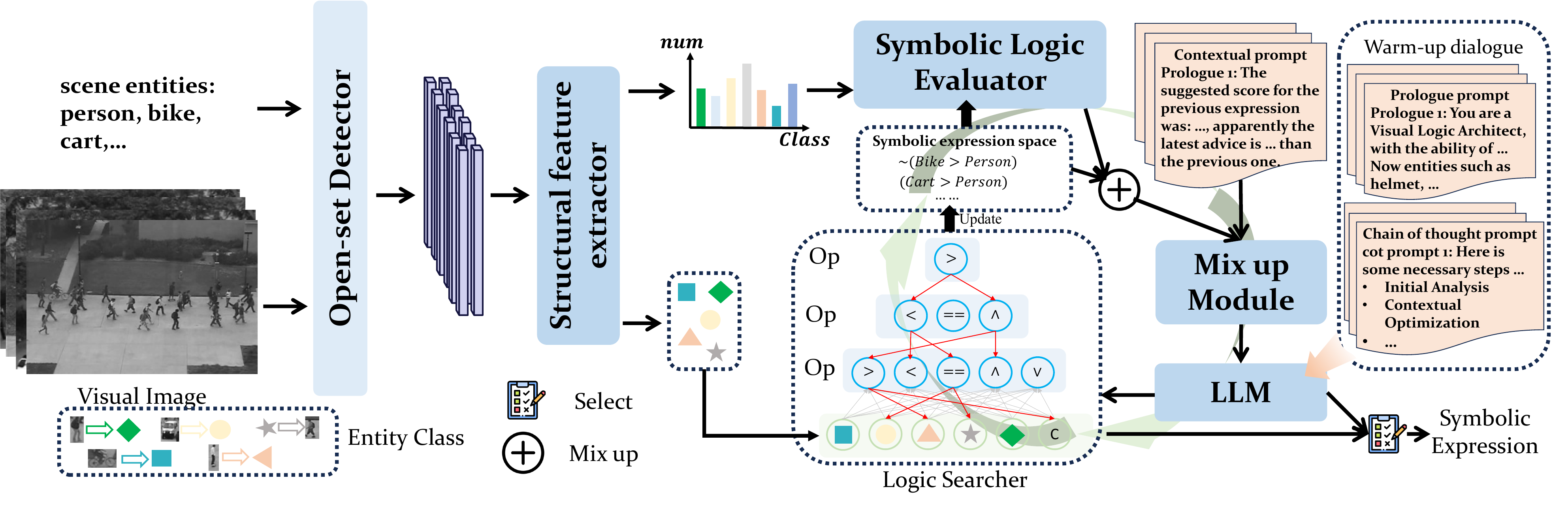}
    \caption{\textbf{Illustration of the proposed \underline{SymbolicDet}}. SymbolicDet mainly consists of logic search and symbolic reasoning. The former module constructs and explores the search space by leveraging structured entity features extracted from an open-set object detector. The latter module harnesses the symbolic reasoning capabilities of Large Language Models (LLMs) along with their inherent commonsense understanding of visual event patterns to guide the search process toward more appropriate and rational pathways.}
    \label{fig:tutorial}
    \vspace{-0.5cm}
\end{figure*}
In practical applications, merely detecting objects often fails to satisfy real-world engineering requirements. Many scenarios demand recognition of complex object relationships or events, which remains a significant challenge in current research. Construction site monitoring requires identifying not just workers and equipment but safety violations; traffic analysis needs to recognize not only vehicles but also dangerous driving patterns; and surveillance systems must detect not merely people but suspicious behaviors. While specialized event recognition systems exist, they typically require extensive training data and lack interpretability.
Our framework addresses this challenge by unlocking the latent event understanding capabilities in standard object detectors through LLM-guided symbolic reasoning. Here, we provide a detailed description of our approach, which transforms object detections into interpretable event recognition without additional training, See appendix for symbol meanings.. Figure \ref{fig:tutorial} illustrates our framework's architecture. \par
Formally, given a visual dataset $\mathcal{D}$ consisting of n images:
\begin{equation}
    \mathcal{D} = \{\left(I_{1},y_{1} \right), \left(I_{2},y_{2} \right), \ldots, \left(I_{n},y_{n} \right)\}
\end{equation}
where each pair consists of an image \(I_i\) and a binary label \(y_i \in \{0, 1\}\) indicating whether a target event \(\varepsilon\) occurs in the image. More specifically, for each image \(I_i\), a standard object detector \(\mathcal{T}\) produces a set of detections \(D_i = \{d_1, d_2, \ldots, d_m\}\), where each detection \(d_j = (c_j, b_j, s_j)\) consists of a category label \(c_j\), a bounding box \(b_j\), and a confidence score \(s_j\). These detections represent ``what exists" in the image.
Our approach seeks to discover an interpretable symbolic expression $f$ that operates solely on the detector outputs to recognize the event. \par
Furthermore, the derived symbolic expression is utilized to assess its capacity in accurately classifying whether the target event is present within an image:
\begin{equation}
    f : \mathcal{O}_I \rightarrow \{0, 1\}\
\end{equation}
This symbolic expression effectively bridges the gap between low-level object detections and high-level event understanding, transforming ``what exists" into ``what is happening" through logical reasoning over detected entities and their relationships. These components operate synergistically, constituting an integrated workflow:
\begin{equation}
    f^* = \arg\max_{f \in \mathcal{F}} \mathcal{G}_{LLM}(f, S(f, \mathcal{T}(\mathcal{D})))
\end{equation}
Where \(f^*\) is the optimal discovered symbolic expression, \(\mathcal{F}\) is the space of all possible expressions in our symbolic language, \(\mathcal{T}\) is the object detector, \(S\) is a scoring function that evaluates how well an expression distinguishes positive and negative examples, and \(\mathcal{G}_{LLM}\) is the LLM guidance mechanism that directs the search toward promising expressions.
In the following sections, we detail each component of our framework and how they work together to unlock event understanding capabilities in standard object detectors.
\subsection{Symbolic Logic Search}
The core of our framework is the symbolic pattern discovery mechanism that identifies meaningful logical expressions capable of recognizing complex events from object detections. This process begins with extracting structured entity representations from detector outputs and then proceeds to search for effective symbolic patterns. \par
\noindent\textbf{Entity-level Feature Extraction.} We first leverage an open-vocabulary object detector to extract comprehensive entity information from each sample. For a given image $I_i$, we obtain a set of entities \(D_i\). These entities form the basis for our symbolic pattern analysis.
To facilitate symbolic reasoning, we transform the raw entity information into structured features:
\begin{equation}
    \mathbf{X} = \{\phi_1(D_1), \phi_2(D_2), ..., \phi_d(D
    _n)\}
\end{equation}
where $\phi_i(\cdot)$ represents different feature extraction functions that capture entity counts, spatial relationships, and attribute distributions. \par
\noindent\textbf{Symbolic Pattern Discovery}
Given the entity representation of images, we next seek to discover symbolic patterns that effectively distinguish images containing the target event from those that do not. The symbolic regression problem is formulated as:
\begin{equation}
    f^* = \arg\min_{f \in \mathcal{F}} \sum_{i=1}^{n} \mathcal{L}(f(\mathbf{X}_i), y_i) + \lambda \Omega(f)
\end{equation}
where $\mathcal{F}$ is the space of possible symbolic expressions, $\mathcal{L}(\cdot)$ is a fitness function measuring pattern discrimination ability, and $\Omega(f)$ is a complexity penalty that promotes simpler expressions.
The search space $\mathcal{F}$ consists of mathematical operators 
$\{+, -, \times, \div, \max, \min\}$ 
and logical operators $\{\land, \lor, \neg\}$. 
To efficiently explore this space, we employ an evolutionary algorithm that initializes a population of candidate expressions, evaluates their fitness on the current dataset, applies genetic operators (mutation, crossover) to generate new candidates, and selects the best expressions for the next generation.
This process generates human-interpretable symbolic expressions that capture meaningful patterns in the data. For example, in a safety helmet detection scenario, a discovered pattern might be:
\begin{equation}
    f(\mathbf{X}) = \bigvee_{i \in \{p,d\}} [\phi_i(E) > \phi_h(E)]
\end{equation}
where $\phi_p$, $\phi_h$, and $\phi_d$ represent the counting functions for persons, helmets, and heads respectively.
While evolutionary search provides a systematic approach to exploring the expression space, its effectiveness is constrained by the stochastic nature of the search process and the exponential growth of the search space with expression complexity. This fundamental challenge highlights the critical role of our LLM guidance mechanism, which strategically directs the evolutionary process toward promising regions of the expression space, balancing exploration with semantic understanding. This synergistic integration, detailed in the following section, enables SymbolicDet to overcome the computational limitations of conventional symbolic approaches while maintaining interpretability.
\subsection{Automated LLM Reasoning}
To enhance the efficiency and effectiveness of symbolic pattern discovery, we propose an automated reasoning mechanism that leverages the semantic understanding capabilities of Large Language Models (LLMs). This LLM-guided approach consists of two main components: a structured prompt space for eliciting effective reasoning and an integrated symbolic search mechanism that combines LLM suggestions with systematic exploration.
\subsubsection{Structured Prompt Space}
We design a hierarchical prompt space to facilitate effective communication with LLMs through three key components:\par
\noindent
\textbf{Scene Context Initialization.} The first layer of prompts establishes the scene context:
\begin{equation}
    P_{\text{init}} = \{\text{scene}, \text{entities}, \text{constraints}\}
\end{equation}
This activation prompt triggers the LLM's prior knowledge relevant to the specific visual event understanding scenario, creating crucial connections between visual entities and semantic understanding.\par
\noindent
\textbf{Chain-of-Thought Guidance.} The second layer provides structured reasoning steps:
\begin{equation}
    P_{\text{cot}} = \{s_1 \rightarrow s_2 \rightarrow ... \rightarrow s_k\}
\end{equation}
where each step $s_i$ guides the LLM through professional analytical frameworks for identifying potential visual event symbolic patterns. This systematic approach ensures comprehensive consideration of entity relationships and domain constraints. \par
\noindent
\textbf{Contextual Feedback Integration.} The final layer incorporates evaluation feedback:
\begin{equation}
    P_{\text{feed}} = \{(r_1, \alpha_1), (r_2, \alpha_2), ..., (r_n, \alpha_n)\}
\end{equation}
where $r_i$ represents previous reasoning attempts and $\alpha_i$ their corresponding effectiveness scores. This feedback mechanism enables the LLM to refine its suggestions based on historical performance.
\begin{algorithm}[t]
\caption{LLM-Guided Symbolic Search}
\begin{algorithmic}[1]
\STATE Initialize population $\mathcal{P}_0$ of symbolic expressions
\FOR{each iteration $t$}
    \STATE $f_t^* \gets$ Select best expression from $\mathcal{P}_t$
    \STATE $S_t \gets$ Generate LLM suggestions via prompt space
    \STATE $\mathcal{P}_{t+1} \gets$ Update population using $\{f_t^*, S_t\}$
    \IF{convergence criterion met}
        \STATE \textbf{break}
    \ENDIF
\ENDFOR
\RETURN best expression $f^*$
\end{algorithmic}
\end{algorithm}
\subsubsection{LLM-Guided Symbolic Search}
We establish a bidirectional interaction mechanism between LLM reasoning and symbolic logic search (SLS) through an iterative process:
At each iteration, the best symbolic expression $f_t^*$ serves as a directional indicator for LLM reasoning. The LLM analyzes this expression through our structured prompt space and generates suggestions $S_t$ that incorporate its semantic understanding and common sense knowledge. These suggestions are then transformed into new symbolic expressions and integrated into the population for the next iteration.
The integration creates a synergistic effect where:
\begin{itemize}
    \item LLM reasoning guides the symbolic search towards semantically meaningful patterns
    \item Symbolic search provides objective evaluation of LLM suggestions
    \item The iterative process combines the interpretability of symbolic expressions with the rich semantic understanding of LLMs
\end{itemize} 
This bidirectional interaction accelerates the discovery of meaningful symbolic patterns while maintaining the interpretability of the detection process. The LLM's suggestions help navigate the vast space of possible symbolic expressions, while the symbolic search framework ensures that the final patterns remain explicit and verifiable. \par
\noindent
\textbf{Analysis.} Through the integration of symbolic logic search and automated LLM reasoning, our framework offers several key advantages for event understanding in object detection: 
First, our approach produces inherently \textit{interpretable results} through complementary mechanisms. The discovered symbolic expressions provide explicit, human-readable logical patterns that directly explain recognition decisions, while the LLM's reasoning process offers semantic context for these patterns. This dual-layer interpretability is critical for applications where understanding the reasoning process is as important as the final decision. 
Second, the bidirectional interaction between symbolic search and LLM reasoning creates an \textit{efficient optimization process}. The LLM's semantic understanding helps navigate the vast space of possible symbolic expressions, while the symbolic search framework grounds the LLM's suggestions in empirical performance. Our structured prompt design ensures systematic utilization of LLM capabilities while maintaining consistency and reproducibility in the reasoning process.
Finally, our framework provides significant \textit{practical deployment advantages}. By operating on the outputs of existing object detectors, it eliminates the need for large-scale training datasets typically required by deep learning methods. This detector-agnostic approach can work with any state-of-the-art detection system without modification, benefiting from advances in object detection while maintaining focus on higher-level event recognition through transparent symbolic reasoning.
\noindent In summary, our framework bridges the gap between low-level object detection and high-level event understanding through a synergistic combination of symbolic search and LLM reasoning. By discovering interpretable symbolic expressions that operate on detector outputs, we unlock event recognition capabilities without extensive training data or specialized architectures, while maintaining full transparency in the reasoning process.
\section{Experiment}
\label{exper}

\begin{table*}[t!]
\caption{Performance of different open set detectors on multiple data sets with or without SymbolicDet module. (AUROC\%) }
\label{table:diff_detector}
\centering
\begin{threeparttable}
\setlength{\tabcolsep}{2.0mm}
\renewcommand\arraystretch{1.3} 
\begin{tabular}{lc|c>{\columncolor{cyan!10}}c|c>{\columncolor{cyan!10}}c|c>{\columncolor{cyan!10}}c}
\toprule
\multicolumn{2}{c|}{Datasets}                         & \multicolumn{2}{c|}{APE \cite{shen2024aligning}}       & \multicolumn{2}{c|}{YOLO-World \cite{cheng2024yolo}} & \multicolumn{2}{c}{GLIP \cite{li2022grounded}}       \\ \cline{3-8} 
                    
                         & \multicolumn{1}{l|}{} & \multicolumn{1}{l}{Original} & +SymbolicDet & \multicolumn{1}{l}{Original}  & +SymbolicDet & \multicolumn{1}{l}{Original} & +SymbolicDet \\ \hline
\multicolumn{1}{l|}{}    & BALL                  & 55.36              & \textbf{94.91} \textcolor{red}{(+39.55)} & 54.76               & \textbf{89.05} \textcolor{red}{(+34.29)} & 66.34              & \textbf{90.27} \textcolor{red}{(+23.93)} \\ \cline{2-8} 
\multicolumn{1}{l|}{ERA \cite{eradataset}} & PersonCrowd           & 78.30              & \textbf{83.26} \textcolor{red}{(+4.96)} & 55.00               & \textbf{85.11} \textcolor{red}{(+30.11)} & 81.71              & \textbf{85.08} \textcolor{red}{(+3.37)} \\ \cline{2-8} 
\multicolumn{1}{l|}{}    & Sport                 & 67.13              & \textbf{90.29} \textcolor{red}{(+23.16)} & 67.27               & \textbf{88.54} \textcolor{red}{(+21.27)} & 66.94              & \textbf{89.65} \textcolor{red}{(+22.71)} \\ \hline
\multicolumn{2}{c|}{Helmet-Mac}                      & 67.41              & \textbf{83.18} \textcolor{red}{(+15.77)} & 65.40               & \textbf{82.47} \textcolor{red}{(+17.07)} & 61.06              & \textbf{76.25} \textcolor{red}{(+15.19)} \\ 
\multicolumn{2}{c|}{Multi-rods Fishing\tnote{1} }                    & 66.82              & \textbf{75.16} \textcolor{red}{(+8.36)} & 52.72               & \textbf{72.01} \textcolor{red}{(+19.29)} & 50.00              & \textbf{71.11} \textcolor{red}{(+21.11)} \\ \hline
\bottomrule
\end{tabular}
\begin{tablenotes}
\item[1] It refers to a subset of Multi-Event Dataset.
\end{tablenotes}
\end{threeparttable}
\end{table*}
\subsection{Experimental Setup}
To comprehensively evaluate our approach, we conduct experiments across diverse datasets spanning various event detection scenarios, comparing against both traditional methods and architecture variants to demonstrate the efficacy of our LLM-guided symbolic reasoning framework.
\subsubsection{Self-collected Datasets}
\noindent
\textbf{Multi-Event Dataset} is our large-scale collection containing over 110,000 images spanning various event detection scenarios. The dataset comprises over 110,000 images spanning various scenarios including fires and waste incineration (30,954 images), multi-rods fishing (12,000 images), night fishing (97 images), license plate detection (12,487 images), personnel loitering and intrusion (10,788 images), among others. For this study, we specifically focus on the multiple-rod fishing scenario, where we have 15,000 training images with 45,341 detailed bounding box annotations covering persons, fishing rods, tackle bags, and umbrellas. The test set contains 2,283 images, with 1,098 cases showing anomalous multi-rods fishing activities and 1,185 normal cases with single-rod fishing. \par
\noindent
\textbf{Helmet-Mac Dataset}, which we make publicly available\footnote{Dataset will soon be available in our code repository.}, addresses the critical domain of construction site safety monitoring. This dataset is curated from various construction scenarios and focuses on safety helmet compliance detection. It contains 7,571 training images with detailed annotations of human heads and safety helmets across diverse construction environments. The test set comprises 4,642 images, balanced between 2,276 safety violations (workers without helmets) and 2,366 compliant cases. The dataset captures various challenging scenarios including different lighting conditions, viewing angles, and occlusion cases, making it a valuable benchmark for safety-critical event detection systems.
\subsubsection{Public Benchmarks}
\noindent
\textbf{ERA Dataset} \cite{eradataset} (Event Recognition in Aerial videos) provides a comprehensive collection of aerial footage covering various event categories. We organize our evaluation around three main event categories: BALL events (327 images) encompassing baseball, soccer, and basketball games; Person\_crowded events (352 images) including conflicts, parade protests, and parties; and Sport events (258 images) covering cycling, boating, and racing activities. Additionally, we utilize 347 Non-event images as negative samples, creating a balanced evaluation framework for our method's discriminative capabilities across different event types. \par
\noindent
\textbf{UCSD Ped2 Dataset} \cite{wang2010anomaly} serves as our primary benchmark for comparison with state-of-the-art methods. This well-established dataset has been widely used in the anomaly detection community, providing a standardized evaluation platform. We use this dataset to demonstrate our method's competitive performance against existing approaches while maintaining the advantages of training-free operation and interpretability.

\subsubsection{Implementation Details}
Our implementation integrates three key components: open-vocabulary object detection, LLM-guided symbolic pattern discovery, and Computational Resources. \par
\noindent
\textbf{Open-vocabulary object detector Setup}
We employ three state-of-the-art multimodal detectors in our experiments: APE \cite{shen2024aligning} (serving as the primary detector), GLIP \cite{li2022grounded}, and YOLO-WORLD \cite{cheng2024yolo}. To optimize detection performance, we implement a two-stage prompt generation process. Initially, we leverage LLM to analyze event scenario descriptions and generate comprehensive detection prompts. The LLM generates prompts not only for objects directly associated with event scenarios but also for contextually related non-event objects, ensuring comprehensive coverage of potential scene elements. Based on empirical studies of detector characteristics, we configure different detection thresholds for optimal performance. Considering APE's characteristically lower threshold nature, we set its minimum detection threshold to 0.05. For GLIP and YOLO-WORLD, we establish a higher threshold of 0.1 to maintain a balance between precision and recall in object detection.  \\
\noindent
\textbf{Symbolic Regression Configuration}
The symbolic regression module processes the detection results through an iterative optimization procedure. Upon receiving detection outputs, the module generates initial logical expressions and evaluates their fitness. If termination criteria are not met, it selects the top-4 logical expressions for crossover mutation, continuing this process until reaching optimal expressions.
We configure the symbolic regression parameters based on extensive experimental validation. The population size is set to twice the number of target categories, allowing for sufficient expression diversity. The crossover and mutation factors are set to 0.5 and 0.3 respectively, providing a balanced exploration-exploitation trade-off. The optimization process continues for 5,000 iterations or until convergence criteria are satisfied. \\
\noindent
\textbf{Computational Resources}
Our experimental framework utilizes a mixed compute infrastructure optimized for different computational demands. Object detection inference is performed on a single NVIDIA RTX 4090 GPU with 24GB memory. Due to our framework's plug-and-play design, even traditional object detectors can be integrated with minimal resource requirements, making our approach adaptable to various hardware configurations. The symbolic regression component of SymbolicDet runs on Intel(R) Xeon(R) Silver 4214R CPU processors, which are well-suited for the parallel exploration of symbolic search. For LLM reasoning, we utilize qwen-series models. This distributed computational approach ensures efficient processing of our training-free pattern discovery pipeline while maintaining practical performance for real-world applications.

\subsection{Main Results}
We evaluate our approach from multiple perspectives: effectiveness across different object detection architectures, comparison with fine-tuning approaches, and benchmarking against traditional event detection methods. \par
\noindent
\textbf{Comparison with Different Detection Architectures.} We first evaluate our framework using three state-of-the-art open-vocabulary detectors: APE, GLIP, and YOLO-WORLD, which represent diverse architectural choices in both visual and language processing. These detectors employ different visual backbones (VIT \cite{dosovitskiy2020image}, Swin-L \cite{liu2021swin}, and YOLOv8) and language models (EVA-CLIP \cite{sun2023eva}, CLIP \cite{radford2021learning}, and BERT \cite{devlin2019bert}). As shown in Table \ref{table:diff_detector}, all three detectors achieve strong performance without any fine-tuning, with APE consistently outperforming others across all five anomaly event scenarios. This superior performance of APE can be attributed to its more sophisticated visual-language alignment mechanism and larger pre-training dataset, which enables better transfer of knowledge to anomaly detection tasks. The consistent performance across architecturally diverse models also suggests that our framework's effectiveness is not tied to specific architectural choices, but rather stems from the fundamental synergy between symbolic reasoning and detection capabilities. \par
\noindent
\textit{Finding 1: The effectiveness of our training-free framework is architecture-agnostic, with APE's superior performance likely due to its enhanced visual-language alignment and broader pre-training.} \par
\noindent
\textbf{Comparison with Fine-tuning Approaches.} To further validate the efficiency of our training-free approach, we conduct comparative experiments with fine-tuned variants on the Helmet-Mac and Multi-rods Fishing datasets. We implement two common fine-tuning strategies: LORA fine-tuning\cite{hu2021lora} and Prompt tuning \cite{lester2021power}, representing different levels of parameter adaptation. Results in Table \ref{table:fine_tune_perf} demonstrate that our training-free approach achieves comparable performance to fine-tuned models. This intriguing finding suggests that our training-free approach effectively leverages the model's general understanding of visual-language relationships. Additionally, the symbolic reasoning component provides a more structured way to capture visual event patterns compared to implicit learning through fine-tuning, achieving similar effectiveness without the computational overhead of parameter adaptation.\par
\noindent
\textit{Finding 2: Our training-free approach is comparable to fine-tuning methods, possibly due to better preservation of general visual-language understanding and more structured pattern discovery.} \par
\begin{table}[!t]
\centering 
\setlength{\tabcolsep}{0.9mm} 
\renewcommand\arraystretch{0.95} 
\caption{Performance of different fine-tuned methods and with or without SymbolicDet module. Lora indicates whether to Lora-tuning the APE model. Prompt indicates whether to Prompt-tuning the APE model (AUROC\%)}
\label{table:fine_tune_perf}
\label{abl_frame}
\begin{tabular}{ccc|c|c}
\toprule
\textbf{Lora} & \textbf{Prompt} & \underline{\textbf{Our}} & \textbf{Helmet-Mac} & \textbf{Multi-rods Fishing}   \\ \midrule
\underline{\hspace{0.2cm}} & \underline{\hspace{0.2cm}}  & \underline{\hspace{0.2cm}}  & 67.41 & 66.82     \\ 
\checkmark & \underline{\hspace{0.2cm}}  & \underline{\hspace{0.2cm}}  & 84.11 & 75.86     \\ 
\underline{\hspace{0.2cm}}  & \checkmark & \underline{\hspace{0.2cm}}  & 67.42 & 66.82     \\ 
\rowcolor{cyan!10} \underline{\hspace{0.2cm}} & \underline{\hspace{0.2cm}} & \checkmark  & 83.18 \small\color{red}(+15.77) & 75.16 \small\color{red}(+8.34)    \\ 
\rowcolor{cyan!10} \checkmark & \underline{\hspace{0.2cm}} & \checkmark  & 95.67 \small\color{red}(+11.56) & 78.44 \small\color{red}(+2.58)    \\ 
\rowcolor{cyan!10} \underline{\hspace{0.2cm}}  & \checkmark &  \checkmark  & 81.62 \small\color{red}(+14.2) & 76.06 \small\color{red}(+9.24)     \\ 
\bottomrule
\end{tabular}
\vspace{-0.5em}
\end{table}
\noindent
\textbf{Comparison with present Methods.} To contextualize our approach within the broader landscape of event detection methods, we evaluate on the UCSD Ped2 and USED\cite{10.1145/2910017.2910624} benchmark and compare against state-of-the-art approaches. As showed in Table \ref{table:ucsd_perf}, our method achieves an impressive 98.7\% accuracy without utilizing training is very close to the current SOTA in UCSD Ped2 dataset and outperforms the baseline in USED dataset. This minimal performance gap is particularly interesting considering the vast difference in approach complexity. We hypothesize that this effectiveness stems from two factors: first, the pre-trained detectors already possess rich semantic understanding that generalizes well to visual event pattern; second, our symbolic reasoning framework effectively translates this semantic knowledge into explicit detection rules, potentially capturing patterns that are similar to those learned by supervised methods but in a more interpretable manner.\par
\noindent
\textit{Finding 3: The near-SOTA performance on UCSD Ped2 suggests that combining pre-trained knowledge with symbolic reasoning can effectively match supervised learning capabilities.} \par
\begin{table}[!t]
\centering 
\setlength{\tabcolsep}{1.5mm} 
\renewcommand\arraystretch{0.95} 
\caption{The overall performance on the UCSD ped2 \cite{wang2010anomaly} and USED\cite{10.1145/2910017.2910624} benchmark.}
\label{table:ucsd_perf}
\label{tab_mlvu}
\begin{tabular}{c|c|c}
\toprule
\textbf{Training-free} & \textbf{Methods}   & \textbf{score (\%) }  \\ \midrule
  & SD-MAE \cite{ristea2024self} & 95.4   \\
 & FastAno \cite{park2022fastano} & 99.3    \\
\textbf{$\times$} & VALD-GAN \cite{singh2024vald}  & 97.74 \\
 & MAMA \cite{hong2024making}  & 98.2  \\
 & Backgroud-Agnostic \cite{georgescu2021background}  & 98.7  \\ 
 & DMAD \cite{liu2023diversity} & \textbf{99.7}  \\ \midrule
\rowcolor{cyan!10} \checkmark & \underline{SymbolicDet}  & \textbf{98.7}   \\
\bottomrule
\end{tabular}
\vspace{-0.5em}
\end{table}
\noindent
\textbf{Ablation Studies.}
We conducted comprehensive ablation studies to evaluate the contribution of each component in our framework. Our analyses examined: (\romannumeral 1) the individual impacts of symbolic regression and manual logic, revealing an 18.36\% performance improvement with symbolic pattern discovery; (\romannumeral 2) the significant benefits of LLM integration on both accuracy and convergence efficiency; and (\romannumeral 3) the positive correlation between search scale and detection performance across datasets. These experiments not only validate SymbolicDet's architectural choices but also confirm that its effectiveness stems from the synergistic combination of symbolic reasoning capabilities and LLM-guided semantic understanding. Detailed results, additional visualizations, and in-depth discussion of these ablation studies are provided in the supplementary materials.
\section{Conclusion}
\label{con}
%
%
In this paper, we introduce \underline{SymbolicDet}, a framework that unlocks event understanding capabilities within standard object detectors through LLM-guided symbolic reasoning. Our approach demonstrates that object detectors contain sufficient visual information for complex event understanding when enhanced with appropriate reasoning mechanisms.
Our key contributions include: First, establishing a paradigm that bridges visual perception and symbolic reasoning through evolutionary pattern discovery and LLM guidance, achieving competitive performance with interpretable reasoning. Second, demonstrating an effective training-free framework that eliminates task-specific fine-tuning. Third, contributing two new benchmark datasets for visual event detection research.
Our results show that combining pre-trained detectors with explicit symbolic reasoning offers a powerful alternative to specialized, training-intensive approaches while enhancing interpretability and adaptability through human-readable symbolic expressions.
Looking forward, while demonstrated in event detection, our approach of enhancing pre-trained visual models with explicit reasoning has broader potential. Future work could extend this framework to relationship detection, behavioral analysis, and intention recognition --- further bridging the gap between perception and reasoning in visual understanding systems.

\section*{ACKNOWLEDGMENTS}
This work was supported by the National Science Fund for Distinguished Young Scholars (No.62025603), the National Natural Science Foundation of China (No. U21B2037, No. U22B2051, No. U23A20383, No. U21A20472, No. 62176222, No. 62176223, No. 62176226, No. 62072386, No. 62072387, No. 62072389, No. 62002305, No. 62272401 and No. 62402015), the Natural Science Foundation of Fujian Province of China (No. 2021J06003, No.2022J06001), the Postdocotoral Fellowship Program of CPSF (GZB20230024) and the China Postdoctoral Science Foundation (2024M750100).

{
    \small
    \bibliographystyle{ieeenat_fullname}
    \bibliography{main}

\begin{thebibliography}{71}
\providecommand{\natexlab}[1]{#1}
\providecommand{\url}[1]{\texttt{#1}}
\expandafter\ifx\csname urlstyle\endcsname\relax
  \providecommand{\doi}[1]{doi: #1}\else
  \providecommand{\doi}{doi: \begingroup \urlstyle{rm}\Url}\fi

\bibitem[Ahmad et~al.(2016)Ahmad, Conci, Boato, and De~Natale]{10.1145/2910017.2910624}
Kashif Ahmad, Nicola Conci, Giulia Boato, and Francesco G.~B. De~Natale.
\newblock Used: a large-scale social event detection dataset.
\newblock In \emph{Proceedings of the 7th International Conference on Multimedia Systems}, New York, NY, USA, 2016. Association for Computing Machinery.

\bibitem[Amizadeh et~al.(2020)Amizadeh, Palangi, Polozov, Huang, and Koishida]{amizadeh2020neuro}
Saeed Amizadeh, Hamid Palangi, Alex Polozov, Yichen Huang, and Kazuhito Koishida.
\newblock Neuro-symbolic visual reasoning: Disentangling.
\newblock In \emph{International Conference on Machine Learning}, pages 279--290. Pmlr, 2020.

\bibitem[Cai and Vasconcelos(2018)]{cai2018cascade}
Zhaowei Cai and Nuno Vasconcelos.
\newblock Cascade r-cnn: Delving into high quality object detection.
\newblock In \emph{Proceedings of the IEEE conference on computer vision and pattern recognition}, pages 6154--6162, 2018.

\bibitem[Carion et~al.(2020)Carion, Massa, Synnaeve, Usunier, Kirillov, and Zagoruyko]{carion2020end}
Nicolas Carion, Francisco Massa, Gabriel Synnaeve, Nicolas Usunier, Alexander Kirillov, and Sergey Zagoruyko.
\newblock End-to-end object detection with transformers.
\newblock In \emph{European conference on computer vision}, pages 213--229. Springer, 2020.

\bibitem[Chen et~al.(2024)Chen, Xu, Kirmani, Ichter, Sadigh, Guibas, and Xia]{chen2024spatialvlm}
Boyuan Chen, Zhuo Xu, Sean Kirmani, Brain Ichter, Dorsa Sadigh, Leonidas Guibas, and Fei Xia.
\newblock Spatialvlm: Endowing vision-language models with spatial reasoning capabilities.
\newblock In \emph{Proceedings of the IEEE/CVF Conference on Computer Vision and Pattern Recognition}, pages 14455--14465, 2024.

\bibitem[Chen and Wu(2024)]{chen2024vtqa}
Kang Chen and Xiangqian Wu.
\newblock Vtqa: Visual text question answering via entity alignment and cross-media reasoning.
\newblock In \emph{Proceedings of the IEEE/CVF Conference on Computer Vision and Pattern Recognition}, pages 27218--27227, 2024.

\bibitem[Cheng et~al.(2024)Cheng, Song, Ge, Liu, Wang, and Shan]{cheng2024yolo}
Tianheng Cheng, Lin Song, Yixiao Ge, Wenyu Liu, Xinggang Wang, and Ying Shan.
\newblock Yolo-world: Real-time open-vocabulary object detection.
\newblock In \emph{Proceedings of the IEEE/CVF Conference on Computer Vision and Pattern Recognition}, pages 16901--16911, 2024.

\bibitem[Devlin et~al.(2019)Devlin, Chang, Lee, and Toutanova]{devlin2019bert}
Jacob Devlin, Ming-Wei Chang, Kenton Lee, and Kristina Toutanova.
\newblock Bert: Pre-training of deep bidirectional transformers for language understanding.
\newblock In \emph{Proceedings of the 2019 conference of the North American chapter of the association for computational linguistics: human language technologies, volume 1 (long and short papers)}, pages 4171--4186, 2019.

\bibitem[Dosovitskiy et~al.(2020)Dosovitskiy, Beyer, Kolesnikov, Weissenborn, Zhai, Unterthiner, Dehghani, Minderer, Heigold, Gelly, et~al.]{dosovitskiy2020image}
Alexey Dosovitskiy, Lucas Beyer, Alexander Kolesnikov, Dirk Weissenborn, Xiaohua Zhai, Thomas Unterthiner, Mostafa Dehghani, Matthias Minderer, Georg Heigold, Sylvain Gelly, et~al.
\newblock An image is worth 16x16 words: Transformers for image recognition at scale.
\newblock \emph{arXiv preprint arXiv:2010.11929}, 2020.

\bibitem[Fan et~al.(2023)Fan, Liu, Li, Wu, and Hua]{fan2023flexible}
Lei Fan, Bo Liu, Haoxiang Li, Ying Wu, and Gang Hua.
\newblock Flexible visual recognition by evidential modeling of confusion and ignorance.
\newblock In \emph{Proceedings of the IEEE/CVF International Conference on Computer Vision}, pages 1338--1347, 2023.

\bibitem[Ganz et~al.(2024)Ganz, Kittenplon, Aberdam, Ben~Avraham, Nuriel, Mazor, and Litman]{ganz2024question}
Roy Ganz, Yair Kittenplon, Aviad Aberdam, Elad Ben~Avraham, Oren Nuriel, Shai Mazor, and Ron Litman.
\newblock Question aware vision transformer for multimodal reasoning.
\newblock In \emph{Proceedings of the IEEE/CVF Conference on Computer Vision and Pattern Recognition}, pages 13861--13871, 2024.

\bibitem[Garcez et~al.(2019)Garcez, Gori, Lamb, Serafini, Spranger, and Tran]{garcez2019neural}
Artur~d'Avila Garcez, Marco Gori, Luis~C Lamb, Luciano Serafini, Michael Spranger, and Son~N Tran.
\newblock Neural-symbolic computing: An effective methodology for principled integration of machine learning and reasoning.
\newblock \emph{arXiv preprint arXiv:1905.06088}, 2019.

\bibitem[Gehrig and Scaramuzza(2023)]{gehrig2023recurrent}
Mathias Gehrig and Davide Scaramuzza.
\newblock Recurrent vision transformers for object detection with event cameras.
\newblock In \emph{Proceedings of the IEEE/CVF conference on computer vision and pattern recognition}, pages 13884--13893, 2023.

\bibitem[Georgescu et~al.(2021)Georgescu, Ionescu, Khan, Popescu, and Shah]{georgescu2021background}
Mariana~Iuliana Georgescu, Radu~Tudor Ionescu, Fahad~Shahbaz Khan, Marius Popescu, and Mubarak Shah.
\newblock A background-agnostic framework with adversarial training for abnormal event detection in video.
\newblock \emph{IEEE transactions on pattern analysis and machine intelligence}, 44\penalty0 (9):\penalty0 4505--4523, 2021.

\bibitem[Girshick(2015)]{girshick2015fast}
Ross Girshick.
\newblock Fast r-cnn.
\newblock In \emph{Proceedings of the IEEE international conference on computer vision}, pages 1440--1448, 2015.

\bibitem[Girshick et~al.(2014)Girshick, Donahue, Darrell, and Malik]{girshick2014rich}
Ross Girshick, Jeff Donahue, Trevor Darrell, and Jitendra Malik.
\newblock Rich feature hierarchies for accurate object detection and semantic segmentation.
\newblock In \emph{Proceedings of the IEEE conference on computer vision and pattern recognition}, pages 580--587, 2014.

\bibitem[Guo et~al.(2023)Guo, Wang, Guo, Li, Song, Tan, Liu, Bian, and Yang]{guo2023connecting}
Qingyan Guo, Rui Wang, Junliang Guo, Bei Li, Kaitao Song, Xu Tan, Guoqing Liu, Jiang Bian, and Yujiu Yang.
\newblock Connecting large language models with evolutionary algorithms yields powerful prompt optimizers.
\newblock \emph{arXiv preprint arXiv:2309.08532}, 2023.

\bibitem[Gupta et~al.(2022)Gupta, Narayan, Joseph, Khan, Khan, and Shah]{gupta2022ow}
Akshita Gupta, Sanath Narayan, KJ Joseph, Salman Khan, Fahad~Shahbaz Khan, and Mubarak Shah.
\newblock Ow-detr: Open-world detection transformer.
\newblock In \emph{Proceedings of the IEEE/CVF conference on computer vision and pattern recognition}, pages 9235--9244, 2022.

\bibitem[Han et~al.(2023)Han, Zhuo, Liao, and Liu]{han2023llms}
Songhao Han, Le Zhuo, Yue Liao, and Si Liu.
\newblock Llms as visual explainers: Advancing image classification with evolving visual descriptions.
\newblock \emph{arXiv preprint arXiv:2311.11904}, 2023.

\bibitem[Hong et~al.(2024)Hong, Ahn, Jo, and Park]{hong2024making}
Seungkyun Hong, Sunghyun Ahn, Youngwan Jo, and Sanghyun Park.
\newblock Making anomalies more anomalous: Video anomaly detection using a novel generator and destroyer.
\newblock \emph{IEEE Access}, 2024.

\bibitem[Hu et~al.(2021)Hu, Shen, Wallis, Allen-Zhu, Li, Wang, Wang, and Chen]{hu2021lora}
Edward~J Hu, Yelong Shen, Phillip Wallis, Zeyuan Allen-Zhu, Yuanzhi Li, Shean Wang, Lu Wang, and Weizhu Chen.
\newblock Lora: Low-rank adaptation of large language models.
\newblock \emph{arXiv preprint arXiv:2106.09685}, 2021.

\bibitem[Hudson and Manning(2019)]{hudson2019gqa}
Drew~A Hudson and Christopher~D Manning.
\newblock Gqa: A new dataset for real-world visual reasoning and compositional question answering.
\newblock In \emph{Proceedings of the IEEE/CVF conference on computer vision and pattern recognition}, pages 6700--6709, 2019.

\bibitem[Jia et~al.(2023)Jia, Yuan, He, Wu, Yu, Lin, Sun, Zhang, and Hu]{jia2023detrs}
Ding Jia, Yuhui Yuan, Haodi He, Xiaopei Wu, Haojun Yu, Weihong Lin, Lei Sun, Chao Zhang, and Han Hu.
\newblock Detrs with hybrid matching.
\newblock In \emph{Proceedings of the IEEE/CVF conference on computer vision and pattern recognition}, pages 19702--19712, 2023.

\bibitem[Khan and Fu(2024)]{khan2024consistency}
Zaid Khan and Yun Fu.
\newblock Consistency and uncertainty: Identifying unreliable responses from black-box vision-language models for selective visual question answering.
\newblock In \emph{Proceedings of the IEEE/CVF Conference on Computer Vision and Pattern Recognition}, pages 10854--10863, 2024.

\bibitem[Krishna et~al.(2017)Krishna, Zhu, Groth, Johnson, Hata, Kravitz, Chen, Kalantidis, Li, Shamma, et~al.]{krishna2017visual}
Ranjay Krishna, Yuke Zhu, Oliver Groth, Justin Johnson, Kenji Hata, Joshua Kravitz, Stephanie Chen, Yannis Kalantidis, Li-Jia Li, David~A Shamma, et~al.
\newblock Visual genome: Connecting language and vision using crowdsourced dense image annotations.
\newblock \emph{International journal of computer vision}, 123:\penalty0 32--73, 2017.

\bibitem[Lester et~al.(2021)Lester, Al-Rfou, and Constant]{lester2021power}
Brian Lester, Rami Al-Rfou, and Noah Constant.
\newblock The power of scale for parameter-efficient prompt tuning.
\newblock \emph{arXiv preprint arXiv:2104.08691}, 2021.

\bibitem[Li et~al.(2023{\natexlab{a}})Li, Tian, and Li]{li2023sodformer}
Dianze Li, Yonghong Tian, and Jianing Li.
\newblock Sodformer: Streaming object detection with transformer using events and frames.
\newblock \emph{IEEE Transactions on Pattern Analysis and Machine Intelligence}, 45\penalty0 (11):\penalty0 14020--14037, 2023{\natexlab{a}}.

\bibitem[Li et~al.(2022{\natexlab{a}})Li, Li, Zhu, Xiang, Huang, and Tian]{li2022asynchronous}
Jianing Li, Jia Li, Lin Zhu, Xijie Xiang, Tiejun Huang, and Yonghong Tian.
\newblock Asynchronous spatio-temporal memory network for continuous event-based object detection.
\newblock \emph{IEEE Transactions on Image Processing}, 31:\penalty0 2975--2987, 2022{\natexlab{a}}.

\bibitem[Li et~al.(2023{\natexlab{b}})Li, Wei, Wang, and Yang]{li2023neural}
Liulei Li, Jianan Wei, Wenguan Wang, and Yi Yang.
\newblock Neural-logic human-object interaction detection.
\newblock \emph{Advances in Neural Information Processing Systems}, 36:\penalty0 21158--21171, 2023{\natexlab{b}}.

\bibitem[Li et~al.(2022{\natexlab{b}})Li, Zhang, Zhang, Yang, Li, Zhong, Wang, Yuan, Zhang, Hwang, et~al.]{li2022grounded}
Liunian~Harold Li, Pengchuan Zhang, Haotian Zhang, Jianwei Yang, Chunyuan Li, Yiwu Zhong, Lijuan Wang, Lu Yuan, Lei Zhang, Jenq-Neng Hwang, et~al.
\newblock Grounded language-image pre-training.
\newblock In \emph{Proceedings of the IEEE/CVF conference on computer vision and pattern recognition}, pages 10965--10975, 2022{\natexlab{b}}.

\bibitem[Li et~al.(2024)Li, Zhang, Lin, Chen, and He]{li2024pixels}
Rongjie Li, Songyang Zhang, Dahua Lin, Kai Chen, and Xuming He.
\newblock From pixels to graphs: Open-vocabulary scene graph generation with vision-language models.
\newblock In \emph{Proceedings of the IEEE/CVF Conference on Computer Vision and Pattern Recognition}, pages 28076--28086, 2024.

\bibitem[Liang et~al.(2023)Liang, Wu, Dai, Li, Zhao, Zhang, Zhang, Vajda, and Marculescu]{liang2023open}
Feng Liang, Bichen Wu, Xiaoliang Dai, Kunpeng Li, Yinan Zhao, Hang Zhang, Peizhao Zhang, Peter Vajda, and Diana Marculescu.
\newblock Open-vocabulary semantic segmentation with mask-adapted clip.
\newblock In \emph{Proceedings of the IEEE/CVF Conference on Computer Vision and Pattern Recognition}, pages 7061--7070, 2023.

\bibitem[Lin et~al.(2021)Lin, Zhao, and Yang]{lin2021complex}
Huan Lin, Hongtian Zhao, and Hua Yang.
\newblock Complex event recognition via spatial-temporal relation graph reasoning.
\newblock In \emph{2021 International Conference on Visual Communications and Image Processing (VCIP)}, pages 1--5. IEEE, 2021.

\bibitem[Lin et~al.(2022)Lin, Ding, Zhang, Zhan, and Tao]{lin2022ru}
Xin Lin, Changxing Ding, Jing Zhang, Yibing Zhan, and Dacheng Tao.
\newblock Ru-net: Regularized unrolling network for scene graph generation.
\newblock In \emph{Proceedings of the IEEE/CVF Conference on Computer Vision and Pattern Recognition}, pages 19457--19466, 2022.

\bibitem[Liu et~al.(2023{\natexlab{a}})Liu, Xu, Yang, Yu, and Yu]{liu2023motion}
Bingde Liu, Chang Xu, Wen Yang, Huai Yu, and Lei Yu.
\newblock Motion robust high-speed light-weighted object detection with event camera.
\newblock \emph{IEEE Transactions on Instrumentation and Measurement}, 72:\penalty0 1--13, 2023{\natexlab{a}}.

\bibitem[Liu et~al.(2023{\natexlab{b}})Liu, Li, Wu, and Lee]{liu2023visual}
Haotian Liu, Chunyuan Li, Qingyang Wu, and Yong~Jae Lee.
\newblock Visual instruction tuning.
\newblock \emph{Advances in neural information processing systems}, 36:\penalty0 34892--34916, 2023{\natexlab{b}}.

\bibitem[Liu et~al.(2022)Liu, Li, Zhang, Yang, Qi, Su, Zhu, and Zhang]{liu2022dab}
Shilong Liu, Feng Li, Hao Zhang, Xiao Yang, Xianbiao Qi, Hang Su, Jun Zhu, and Lei Zhang.
\newblock Dab-detr: Dynamic anchor boxes are better queries for detr.
\newblock \emph{arXiv preprint arXiv:2201.12329}, 2022.

\bibitem[Liu et~al.(2024)Liu, Zeng, Ren, Li, Zhang, Yang, Jiang, Li, Yang, Su, et~al.]{liu2024grounding}
Shilong Liu, Zhaoyang Zeng, Tianhe Ren, Feng Li, Hao Zhang, Jie Yang, Qing Jiang, Chunyuan Li, Jianwei Yang, Hang Su, et~al.
\newblock Grounding dino: Marrying dino with grounded pre-training for open-set object detection.
\newblock In \emph{European Conference on Computer Vision}, pages 38--55. Springer, 2024.

\bibitem[Liu et~al.(2023{\natexlab{c}})Liu, Chang, Ma, Shan, and Chen]{liu2023diversity}
Wenrui Liu, Hong Chang, Bingpeng Ma, Shiguang Shan, and Xilin Chen.
\newblock Diversity-measurable anomaly detection.
\newblock In \emph{Proceedings of the IEEE/CVF conference on computer vision and pattern recognition}, pages 12147--12156, 2023{\natexlab{c}}.

\bibitem[Liu et~al.(2021)Liu, Lin, Cao, Hu, Wei, Zhang, Lin, and Guo]{liu2021swin}
Ze Liu, Yutong Lin, Yue Cao, Han Hu, Yixuan Wei, Zheng Zhang, Stephen Lin, and Baining Guo.
\newblock Swin transformer: Hierarchical vision transformer using shifted windows.
\newblock In \emph{Proceedings of the IEEE/CVF international conference on computer vision}, pages 10012--10022, 2021.

\bibitem[Majumdar et~al.(2024)Majumdar, Ajay, Zhang, Putta, Yenamandra, Henaff, Silwal, Mcvay, Maksymets, Arnaud, et~al.]{majumdar2024openeqa}
Arjun Majumdar, Anurag Ajay, Xiaohan Zhang, Pranav Putta, Sriram Yenamandra, Mikael Henaff, Sneha Silwal, Paul Mcvay, Oleksandr Maksymets, Sergio Arnaud, et~al.
\newblock Openeqa: Embodied question answering in the era of foundation models.
\newblock In \emph{Proceedings of the IEEE/CVF conference on computer vision and pattern recognition}, pages 16488--16498, 2024.

\bibitem[Mao et~al.(2019)Mao, Gan, Kohli, Tenenbaum, and Wu]{mao2019neuro}
Jiayuan Mao, Chuang Gan, Pushmeet Kohli, Joshua~B Tenenbaum, and Jiajun Wu.
\newblock The neuro-symbolic concept learner: Interpreting scenes, words, and sentences from natural supervision.
\newblock \emph{arXiv preprint arXiv:1904.12584}, 2019.

\bibitem[Meng et~al.(2021)Meng, Chen, Fan, Zeng, Li, Yuan, Sun, and Wang]{meng2021conditional}
Depu Meng, Xiaokang Chen, Zejia Fan, Gang Zeng, Houqiang Li, Yuhui Yuan, Lei Sun, and Jingdong Wang.
\newblock Conditional detr for fast training convergence.
\newblock In \emph{Proceedings of the IEEE/CVF international conference on computer vision}, pages 3651--3660, 2021.

\bibitem[Merlo et~al.(2023)Merlo, Lagomarsino, Lamon, and Ajoudani]{merlo2023automatic}
Elena Merlo, Marta Lagomarsino, Edoardo Lamon, and Arash Ajoudani.
\newblock Automatic interaction and activity recognition from videos of human manual demonstrations with application to anomaly detection.
\newblock In \emph{2023 32nd IEEE International Conference on Robot and Human Interactive Communication (RO-MAN)}, pages 1188--1195. IEEE, 2023.

\bibitem[Mou et~al.(in press)Mou, Hua, Jin, and Zhu]{eradataset}
L. Mou, Y. Hua, P. Jin, and X.~X. Zhu.
\newblock {ERA: A dataset and deep learning benchmark for event recognition in aerial videos}.
\newblock \emph{IEEE Geoscience and Remote Sensing Magazine}, in press.

\bibitem[Oquab et~al.(2023)Oquab, Darcet, Moutakanni, Vo, Szafraniec, Khalidov, Fernandez, Haziza, Massa, El-Nouby, et~al.]{oquab2023dinov2}
Maxime Oquab, Timoth{\'e}e Darcet, Th{\'e}o Moutakanni, Huy Vo, Marc Szafraniec, Vasil Khalidov, Pierre Fernandez, Daniel Haziza, Francisco Massa, Alaaeldin El-Nouby, et~al.
\newblock Dinov2: Learning robust visual features without supervision.
\newblock \emph{arXiv preprint arXiv:2304.07193}, 2023.

\bibitem[Pan et~al.(2023)Pan, Xing, Diao, Sun, Liu, Shum, Pi, Zhang, and Zhang]{pan2023plum}
Rui Pan, Shuo Xing, Shizhe Diao, Wenhe Sun, Xiang Liu, Kashun Shum, Renjie Pi, Jipeng Zhang, and Tong Zhang.
\newblock Plum: Prompt learning using metaheuristic.
\newblock \emph{arXiv preprint arXiv:2311.08364}, 2023.

\bibitem[Park et~al.(2022)Park, Cho, Lee, and Lee]{park2022fastano}
Chaewon Park, MyeongAh Cho, Minhyeok Lee, and Sangyoun Lee.
\newblock Fastano: Fast anomaly detection via spatio-temporal patch transformation.
\newblock In \emph{Proceedings of the IEEE/CVF Winter Conference on Applications of Computer Vision}, pages 2249--2259, 2022.

\bibitem[Pi et~al.(2023)Pi, Gao, Diao, Pan, Dong, Zhang, Yao, Han, Xu, Kong, et~al.]{pi2023detgpt}
Renjie Pi, Jiahui Gao, Shizhe Diao, Rui Pan, Hanze Dong, Jipeng Zhang, Lewei Yao, Jianhua Han, Hang Xu, Lingpeng Kong, et~al.
\newblock Detgpt: Detect what you need via reasoning.
\newblock \emph{arXiv preprint arXiv:2305.14167}, 2023.

\bibitem[Radford et~al.(2021)Radford, Kim, Hallacy, Ramesh, Goh, Agarwal, Sastry, Askell, Mishkin, Clark, et~al.]{radford2021learning}
Alec Radford, Jong~Wook Kim, Chris Hallacy, Aditya Ramesh, Gabriel Goh, Sandhini Agarwal, Girish Sastry, Amanda Askell, Pamela Mishkin, Jack Clark, et~al.
\newblock Learning transferable visual models from natural language supervision.
\newblock In \emph{International conference on machine learning}, pages 8748--8763. PmLR, 2021.

\bibitem[Redmon et~al.(2016)Redmon, Divvala, Girshick, and Farhadi]{redmon2016you}
Joseph Redmon, Santosh Divvala, Ross Girshick, and Ali Farhadi.
\newblock You only look once: Unified, real-time object detection.
\newblock In \emph{Proceedings of the IEEE conference on computer vision and pattern recognition}, pages 779--788, 2016.

\bibitem[Ren et~al.(2015)Ren, He, Girshick, and Sun]{ren2015faster}
Shaoqing Ren, Kaiming He, Ross Girshick, and Jian Sun.
\newblock Faster r-cnn: Towards real-time object detection with region proposal networks.
\newblock \emph{Advances in neural information processing systems}, 28, 2015.

\bibitem[Ristea et~al.(2024)Ristea, Croitoru, Ionescu, Popescu, Khan, Shah, et~al.]{ristea2024self}
Nicolae-C Ristea, Florinel-Alin Croitoru, Radu~Tudor Ionescu, Marius Popescu, Fahad~Shahbaz Khan, Mubarak Shah, et~al.
\newblock Self-distilled masked auto-encoders are efficient video anomaly detectors.
\newblock In \emph{Proceedings of the IEEE/CVF Conference on Computer Vision and Pattern Recognition}, pages 15984--15995, 2024.

\bibitem[Saito et~al.(2022)Saito, Hu, Darrell, and Saenko]{saito2022learning}
Kuniaki Saito, Ping Hu, Trevor Darrell, and Kate Saenko.
\newblock Learning to detect every thing in an open world.
\newblock In \emph{European Conference on Computer Vision}, pages 268--284. Springer, 2022.

\bibitem[Sakaino et~al.(2023)Sakaino, Gaviphat, Zamora, Insisiengmay, and Ningrum]{sakaino2023deepunseen}
Hidetomo Sakaino, Natnapat Gaviphat, Louie Zamora, Alivanh Insisiengmay, and Dwi~Fetiria Ningrum.
\newblock Deepunseen: Unpredicted event recognition through integrated vision-language models.
\newblock In \emph{2023 IEEE Conference on Artificial Intelligence (CAI)}, pages 48--50. IEEE, 2023.

\bibitem[Shen et~al.(2024)Shen, Fu, Chen, Zhang, Li, Sun, Wu, Lin, and Ji]{shen2024aligning}
Yunhang Shen, Chaoyou Fu, Peixian Chen, Mengdan Zhang, Ke Li, Xing Sun, Yunsheng Wu, Shaohui Lin, and Rongrong Ji.
\newblock Aligning and prompting everything all at once for universal visual perception.
\newblock In \emph{Proceedings of the IEEE/CVF Conference on Computer Vision and Pattern Recognition}, pages 13193--13203, 2024.

\bibitem[Shi et~al.(2019)Shi, Zhang, and Li]{shi2019explainable}
Jiaxin Shi, Hanwang Zhang, and Juanzi Li.
\newblock Explainable and explicit visual reasoning over scene graphs.
\newblock In \emph{Proceedings of the IEEE/CVF conference on computer vision and pattern recognition}, pages 8376--8384, 2019.

\bibitem[Singh et~al.(2024)Singh, Sethi, Saini, Saurav, Tiwari, and Singh]{singh2024vald}
Rituraj Singh, Anikeit Sethi, Krishanu Saini, Sumeet Saurav, Aruna Tiwari, and Sanjay Singh.
\newblock Vald-gan: video anomaly detection using latent discriminator augmented gan.
\newblock \emph{Signal, Image and Video Processing}, 18\penalty0 (1):\penalty0 821--831, 2024.

\bibitem[Sun et~al.(2023)Sun, Fang, Wu, Wang, and Cao]{sun2023eva}
Quan Sun, Yuxin Fang, Ledell Wu, Xinlong Wang, and Yue Cao.
\newblock Eva-clip: Improved training techniques for clip at scale.
\newblock \emph{arXiv preprint arXiv:2303.15389}, 2023.

\bibitem[Wang et~al.(2023)Wang, Jia, Zhang, Zhang, Wang, Zhang, Wang, and Lu]{wang2023dual}
Dongsheng Wang, Xu Jia, Yang Zhang, Xinyu Zhang, Yaoyuan Wang, Ziyang Zhang, Dong Wang, and Huchuan Lu.
\newblock Dual memory aggregation network for event-based object detection with learnable representation.
\newblock In \emph{Proceedings of the AAAI Conference on Artificial Intelligence}, pages 2492--2500, 2023.

\bibitem[Wang and Miao(2010)]{wang2010anomaly}
Shu Wang and Zhenjiang Miao.
\newblock Anomaly detection in crowd scene.
\newblock In \emph{IEEE 10th International Conference on Signal Processing Proceedings}, pages 1220--1223. IEEE, 2010.

\bibitem[Wang et~al.(2025)Wang, Li, Zhang, Zhang, Xie, Liu, Zeng, and Jin]{wang2025scene}
Yunnan Wang, Ziqiang Li, Wenyao Zhang, Zequn Zhang, Baao Xie, Xihui Liu, Wenjun Zeng, and Xin Jin.
\newblock Scene graph disentanglement and composition for generalizable complex image generation.
\newblock \emph{Advances in Neural Information Processing Systems}, 37:\penalty0 98478--98504, 2025.

\bibitem[Xu et~al.(2022)Xu, Chen, Du, Shao, Wang, Li, and Yang]{xu2022gps}
Hanwei Xu, Yujun Chen, Yulun Du, Nan Shao, Yanggang Wang, Haiyu Li, and Zhilin Yang.
\newblock Gps: Genetic prompt search for efficient few-shot learning.
\newblock \emph{arXiv preprint arXiv:2210.17041}, 2022.

\bibitem[Yi et~al.(2018)Yi, Wu, Gan, Torralba, Kohli, and Tenenbaum]{yi2018neural}
Kexin Yi, Jiajun Wu, Chuang Gan, Antonio Torralba, Pushmeet Kohli, and Josh Tenenbaum.
\newblock Neural-symbolic vqa: Disentangling reasoning from vision and language understanding.
\newblock \emph{Advances in neural information processing systems}, 31, 2018.

\bibitem[Zareian et~al.(2020)Zareian, Karaman, and Chang]{zareian2020bridging}
Alireza Zareian, Svebor Karaman, and Shih-Fu Chang.
\newblock Bridging knowledge graphs to generate scene graphs.
\newblock In \emph{Computer Vision--ECCV 2020: 16th European Conference, Glasgow, UK, August 23--28, 2020, Proceedings, Part XXIII 16}, pages 606--623. Springer, 2020.

\bibitem[Zhang et~al.(2022{\natexlab{a}})Zhang, Li, Liu, Zhang, Su, Zhu, Ni, and Shum]{zhang2022dino}
Hao Zhang, Feng Li, Shilong Liu, Lei Zhang, Hang Su, Jun Zhu, Lionel~M Ni, and Heung-Yeung Shum.
\newblock Dino: Detr with improved denoising anchor boxes for end-to-end object detection.
\newblock \emph{arXiv preprint arXiv:2203.03605}, 2022{\natexlab{a}}.

\bibitem[Zhang et~al.(2022{\natexlab{b}})Zhang, Dong, Zhang, Ding, Heide, Yin, and Yang]{zhang2022spiking}
Jiqing Zhang, Bo Dong, Haiwei Zhang, Jianchuan Ding, Felix Heide, Baocai Yin, and Xin Yang.
\newblock Spiking transformers for event-based single object tracking.
\newblock In \emph{Proceedings of the IEEE/CVF conference on Computer Vision and Pattern Recognition}, pages 8801--8810, 2022{\natexlab{b}}.

\bibitem[Zhang et~al.(2024)Zhang, Ma, Gao, Shakiah, Gao, and Chai]{zhang2024groundhog}
Yichi Zhang, Ziqiao Ma, Xiaofeng Gao, Suhaila Shakiah, Qiaozi Gao, and Joyce Chai.
\newblock Groundhog: Grounding large language models to holistic segmentation.
\newblock In \emph{Proceedings of the IEEE/CVF conference on computer vision and pattern recognition}, pages 14227--14238, 2024.

\bibitem[Zhu et~al.(2020)Zhu, Su, Lu, Li, Wang, and Dai]{zhu2020deformable}
Xizhou Zhu, Weijie Su, Lewei Lu, Bin Li, Xiaogang Wang, and Jifeng Dai.
\newblock Deformable detr: Deformable transformers for end-to-end object detection.
\newblock \emph{arXiv preprint arXiv:2010.04159}, 2020.

\bibitem[Zohar et~al.(2023)Zohar, Wang, and Yeung]{zohar2023prob}
Orr Zohar, Kuan-Chieh Wang, and Serena Yeung.
\newblock Prob: Probabilistic objectness for open world object detection.
\newblock In \emph{Proceedings of the IEEE/CVF Conference on Computer Vision and Pattern Recognition}, pages 11444--11453, 2023.

\bibitem[Zou et~al.(2024)Zou, Yang, Zhang, Li, Li, Wang, Wang, Gao, and Lee]{zou2024segment}
Xueyan Zou, Jianwei Yang, Hao Zhang, Feng Li, Linjie Li, Jianfeng Wang, Lijuan Wang, Jianfeng Gao, and Yong~Jae Lee.
\newblock Segment everything everywhere all at once.
\newblock \emph{Advances in Neural Information Processing Systems}, 36, 2024.

\end{thebibliography}
}
\clearpage
\setcounter{page}{1}
\maketitlesupplementary
\section{Ablation Study}
\noindent
\textbf{Analysis of Component Contributions.} To understand the contribution of each component in our framework, we conduct comprehensive ablation studies examining the individual and combined effects of LLM reasoning and symbolic regression. Starting with a baseline using only manual logic expressions (67.00\% average performance), the addition of symbolic regression significantly improves performance to 85.36\%. This substantial improvement (+18.36\%) suggests that automated pattern discovery through symbolic regression is significantly more effective than human-designed rules, likely due to its ability to explore a broader space of logical combinations and capture subtle patterns that might not be immediately apparent to human experts. \par
\begin{figure}[!t]
    \centering
    \setlength{\abovecaptionskip}{0.1cm}
    \includegraphics[width=1\linewidth]{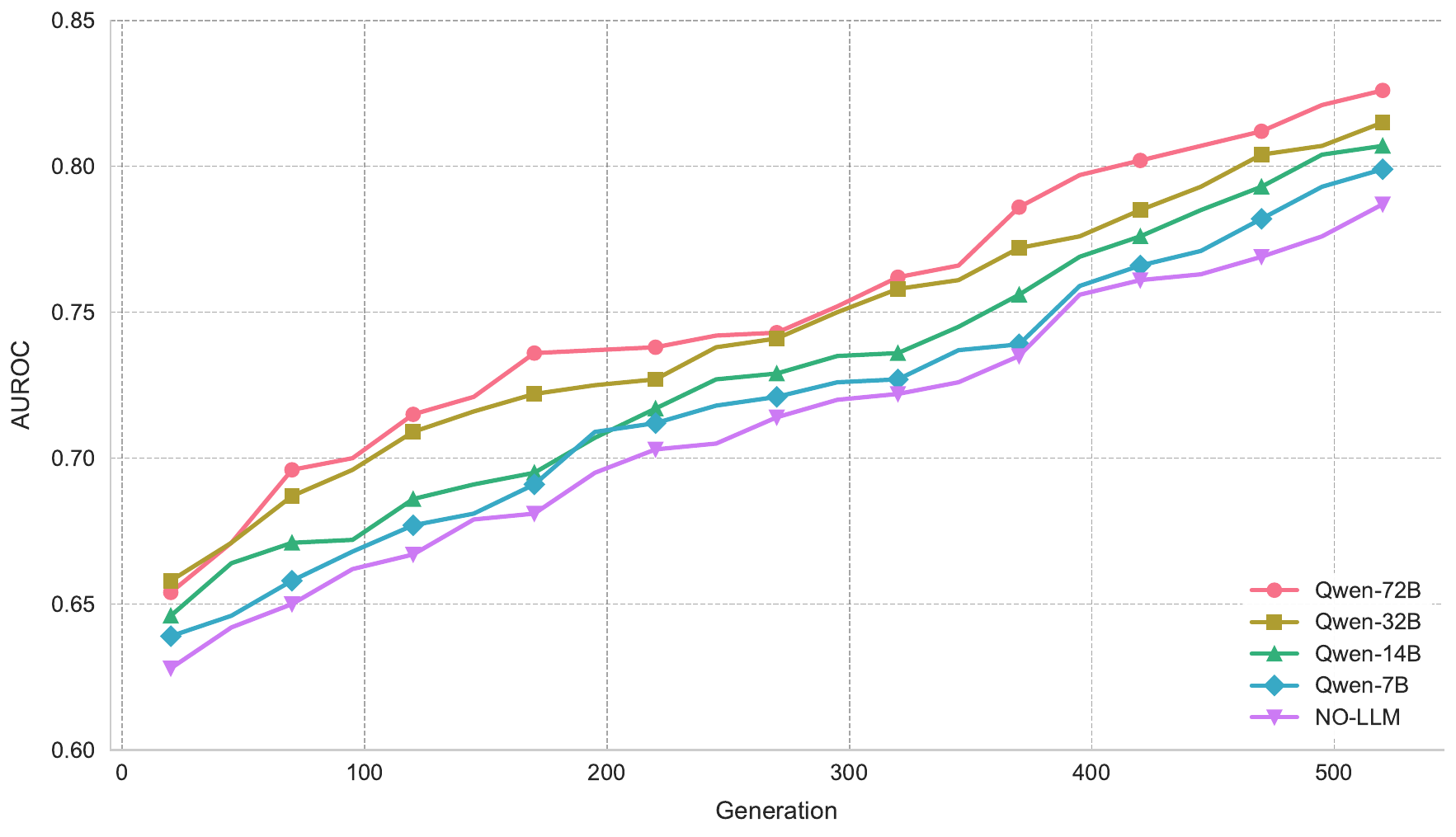}
    \caption{Performance on SymbolicDet with or without LLM.}
    \label{fig:wo_llm}
\end{figure}
\setlength{\columnsep}{10pt}
\begin{table}[!t]
  \setlength{\tabcolsep}{6pt} 
  \begin{tabular}{c|c|c|c|c}
    \toprule
     & \textbf{Ours} & \textbf{NSCL} & \textbf{NS-VQA} & \textbf{LogicHOI} \\
    \midrule
    \textbf{Multi-rods} & 0.75 & 0.65 & 0.61 & 0.67 \\
    \textbf{Helmet} & 0.83 & 0.68 & 0.65 & 0.69 \\
    \bottomrule
  \end{tabular}
  \captionof{table}{ Comparison with other nuero-symbolic methods. (Acc)} 
  \label{table:nscl_perf}
\end{table}
%
%
%
\noindent
\textbf{Impact of LLM Integration.} Figure \ref{fig:wo_llm} illustrates the substantial impact of different LLM integration on both the effectiveness and efficiency of our symbolic pattern discovery process. When examining convergence trajectories across generations, we observe that LLM guidance not only enhances the ultimate detection accuracy but also significantly accelerates the convergence speed of symbolic regression. The analysis compares performance curves with and without LLM guidance, as well as across different LLM scales.
\textit{Finding: Effective event detection through symbolic reasoning benefits from the complementary strengths of systematic pattern discovery (through evolutionary search) and semantic guidance (through LLM reasoning). The symbolic component provides the expressive framework for capturing complex relationships, while the LLM component contributes domain knowledge and conceptual understanding that steers the search toward meaningful patterns.} \par
\begin{figure}[!t]
    \centering
    \setlength{\abovecaptionskip}{0.1cm}
    \includegraphics[width=1\linewidth]{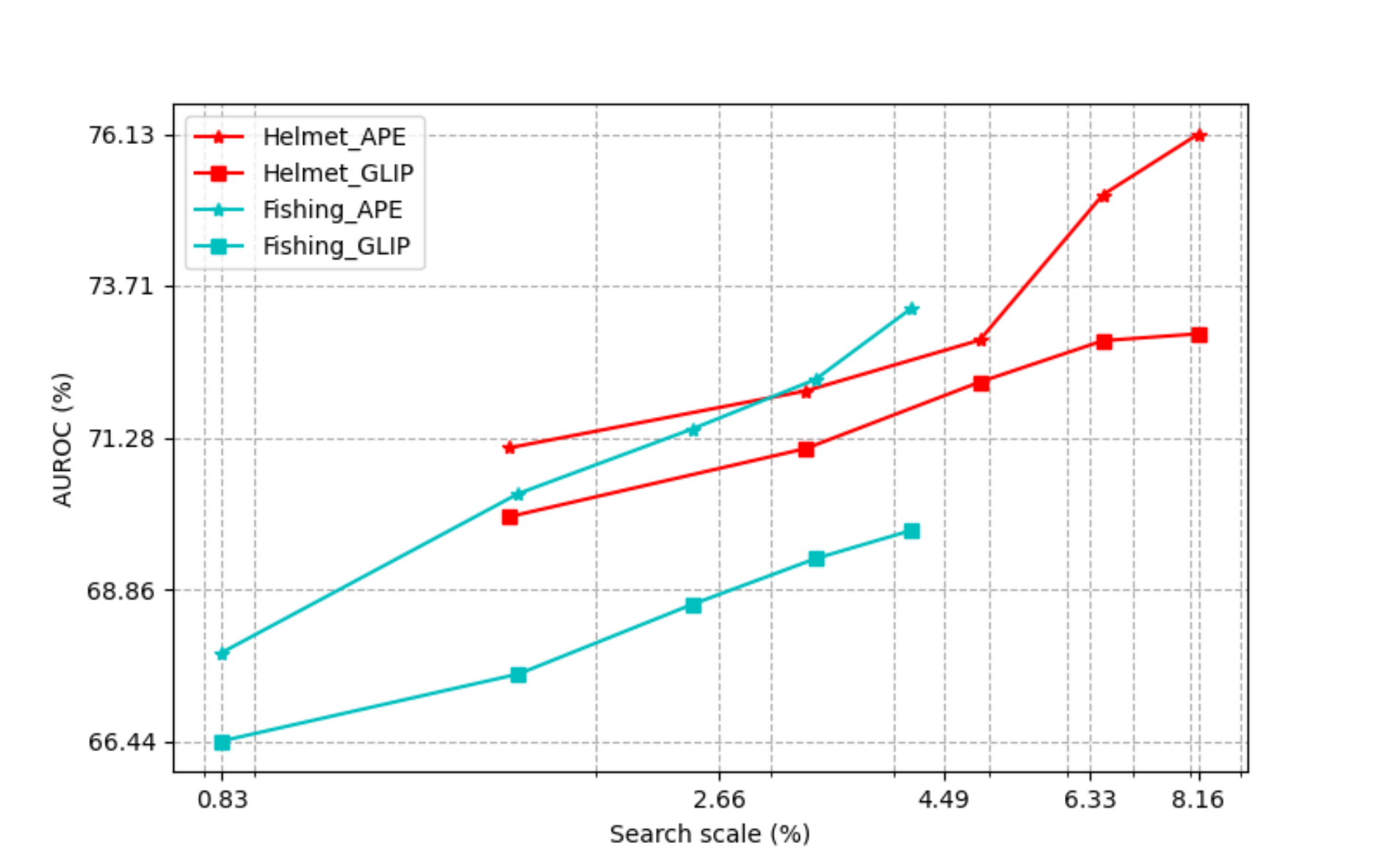}
    \caption{Performance on different search scales.}
    \label{fig:search_scale}
\end{figure}
\begin{table}[!t]
\centering 
\begin{threeparttable}
\setlength{\tabcolsep}{0.9mm} 
\renewcommand\arraystretch{0.8} 
\begin{tabular}{c|c|c|c|c|c}
\toprule
 & \textbf{wo llm} & \textbf{7B} & \textbf{14B} & \textbf{32B} & \textbf{72B}   \\ \midrule
\textbf{Run time(s/500it)} & 80.66  & 265.5  & 281.01 & 268 & 267     \\ 
\textbf{Cost time(s)\tnote{1}} & 69.66  & 45.92  & 44.91 & 30.35 & 26.07 \\ 
\textbf{Memory(MB)} & 293.65  & 218.56  & 218.34 & 218.68 & 218.82 \\ 
\bottomrule
\end{tabular}
\captionof{table}{\centering The computational overhead of symbolic search process.}
\label{table:diff_llm_overhead}
\begin{tablenotes}
\item[1] \small{It refers to the time needed to achieve the same performance.}
\end{tablenotes}
\end{threeparttable}
\end{table}
\noindent
\textbf{Effect of Search Scale on Performance.} To further explore the robustness of our framework, we investigate the effect of varying search scales on event detection accuracy, as depicted in Figure \ref{fig:search_scale}. The search scale defines the proportion of samples allocated for constructing the logical search space, with the remainder used for pattern evaluation.
Our results reveal a clear pattern: increasing the search scale consistently enhances AUROC performance across both the Helmet and Fishing datasets using APE and GLIP strategies. Notably, in the Helmet dataset, both strategies show a significant improvement, reaching peak performance at the highest search scale of 8.16\%. The Fishing dataset demonstrates a similar upward trend, highlighting the benefits of expanding the search space.
\textit{Finding:} The increase in performance with larger search scales underscores the efficacy of our approach in utilizing more extensive logical reasoning. The findings suggest that even without traditional fine-tuning, enlarging the search space enables the framework to uncover more accurate and interpretable patterns. This scalability evidences the flexibility and potency of SymbolicDet in capitalizing on the latent potential of standard object detectors, reinforcing its applicability across diverse scenarios.

\section{Extra experiments}
\noindent
We conducted additional experiments by transferring several representative neuro-symbolic methods to be evaluated on our benchmark dataset. Detailed results can be found in Table \ref{table:nscl_perf}.
\noindent
In addition, we conducted experiments to compare the computational overhead. We analyze the trade-off between computational cost and performance when using LLMs with varying parameter counts versus not using an LLM at all.Details can be found in Table \ref{table:diff_llm_overhead}. 
We also analyze the computational costs of each step of our work.The main computational steps of this work are: Detector inference, Symbolic search process and LLM-guided reasoning. It is worth noting that we use a third-party LLM service provider for the third step of computation, so the actual computational cost may vary significantly depending on the service provider. Detailed computational costs for each part are shown in Table \ref{table:overhead_each_part}.
\par
\begin{table}[!t]
\centering 
\renewcommand\arraystretch{0.5} 
\begin{tabular}{c|c|c|c}
\toprule
 & \textbf{Detection} & \textbf{Search} & \textbf{LLm Infer}  \\ \midrule
\textbf{Run time(s/iter)} & 0.02-1.5  & 0.16  & 8-14  \\ 
\textbf{GPU Memory(GB)} & 0.54-18  & -  & -   \\ 
\bottomrule
\end{tabular}
\captionof{table}{\centering The computational overhead of each part.}
\label{table:overhead_each_part}
\end{table}
\noindent
In addition, we have conducted additional experiments on three subsets of the USED dataset: SPORT, CONCERT, and PROTEST. The results from these new experiments will be presented in Table \ref{table:results_in_used}. \par
\begin{table}[!t]
\centering 
\renewcommand\arraystretch{0.5} 
\begin{tabular}{c|c|c|c}
\toprule
  & \textbf{SPORT} & \textbf{CONCERT} & \textbf{PROTEST}  \\ \midrule
\textbf{Ours}  & 0.93  & 0.99 & 0.92  \\ 
\textbf{USED}  & 0.66 & 0.75 & 0.67\\ 
\bottomrule
\end{tabular}
\captionof{table}{\centering The computational overhead of each part. (Acc)}
\label{table:results_in_used}
\end{table}

\begin{figure*}[!t]
    \centering
    \setlength{\abovecaptionskip}{-0.1cm}
    \includegraphics[width=0.95\linewidth]{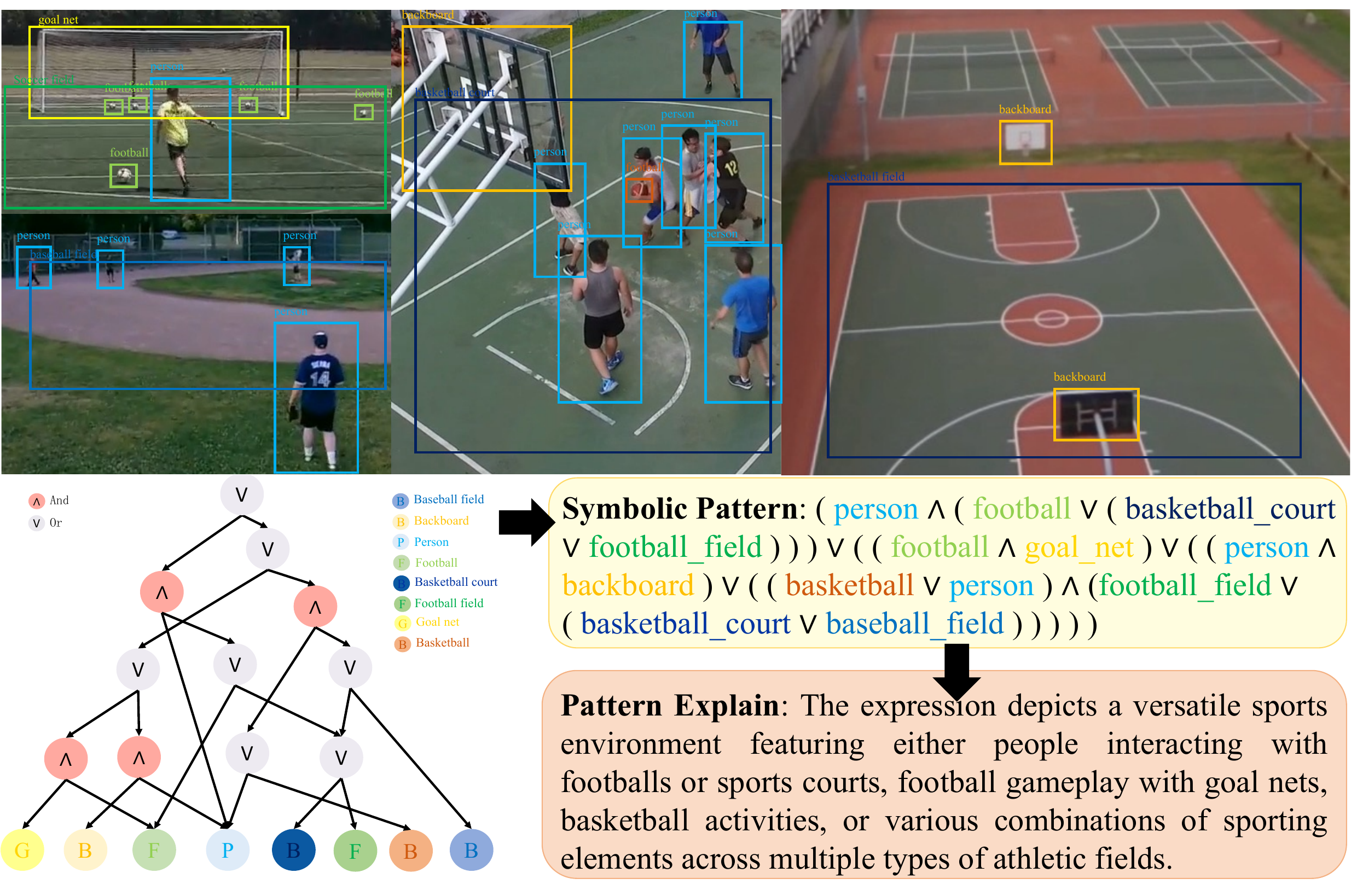}
    \caption{Illustration of the application of symbolic pattern detection in sports environments, showcasing how logical expressions can be used to identify and categorize complex sports scenarios.}
    \label{fig:display_ball_scene}
    \vspace{-0.3cm}
\end{figure*}



\begin{table*}[!ht] 
    \centering 
    \begin{tabular}{cc}
        \toprule 
        \textbf{Symbol}& \textbf{Definition}\\
        \midrule 
        $\mathcal{D}$& A visual datasdet\\
        $I_i$& An image i in visual dataset\\
        $y_i$& A binary label corresponds to image i\\
        ${D}$& A set of objects which are detected by object detector\\
        $d_j$& A detected object \\
        $c_j$& Category label belongs to $d_j$\\
        $b_j$& Bounding box belongs to $d_j$\\
        $s_j$& Confidence score belongs to $d_j$\\
        $\epsilon$& Target event\\
        $f$& Symbolic expression\\
        $\mathcal{O}_I$& A binary symbol to judge whether the target event is present within an image I\\
        $f^*$& Optimal discovered symbolic expression\\
        $\mathcal{F}$& The space of all possible expressions in our symbolic language\\
        $\mathcal{T}$& Object detector\\
        $S$& A scoring function that evaluates how well an expression distinguishes positive and negative examples\\
        $\mathcal{G}_{LLM}$& The LLM guidance mechanism that directs the search toward promising expressions\\
        $\phi_i(\cdot)$& Feature extraction functions that capture entity counts, spatial relationships, and attribute distributions\\
        $\Omega(f)$& A complexity penalty that promotes simpler expressions\\
        $P_{\text{init}}$& Initial prompt for the LLM guidence\\
        $P_{\text{cot}}$& Prompt for chain-of-thought\\
        $P_{\text{feed}}$& Prompt after contextual feedback integration\\
        \bottomrule 
    \end{tabular}
    \caption{Symbol Definition}
    \label{tab:notation}
\end{table*}

\section{Notation and Results}
This section provides a comprehensive list of mathematical symbols and notations used throughout this paper, followed by visual demonstration of our proposed method's performance.
Table \ref{tab:notation} summarizes the key mathematical symbols and their definitions used in this work.
Figure \ref{fig:display_ball_scene} illustrates the effectiveness of our proposed approach through visual comparison and performance metrics.


\end{document}